\pdfoutput=1

\documentclass[11pt]{article}
\usepackage[final]{acl}

\usepackage{times}
\usepackage{latexsym}
\usepackage[T1]{fontenc}
\usepackage[utf8]{inputenc}
\usepackage[ruled,vlined]{algorithm2e}
\usepackage{amsfonts}
\usepackage{amsmath}
\usepackage{amssymb}        
\usepackage{bold-extra}     
\usepackage{bm}             

\usepackage{graphicx}       
\usepackage{multirow}       
\usepackage{booktabs}       
\usepackage{enumitem}       
\usepackage{subcaption}     
\usepackage[hang,flushmargin]{footmisc}  
\usepackage{hyperref}
\title{Modeling Student Performance in Game-Based Learning Environments}
\author{
Hyunbae Jeon\thanks{These authors contributed equally to this work.} \\
  Department of Computer Science\\
  Emory University \\
  Atlanta, GA, 30322, USA \\
  \texttt{hjeon62@emory.edu} \\\And
  Harry He\footnotemark[1] \\
  Department of Computer Science\\
  Emory University \\
  Atlanta, GA, 30322, USA \\
  \texttt{hhe48@emory.edu} \\\AND
  Anthony Wang\footnotemark[1] \\
  Department of Biology \\
  Emory University \\
  Atlanta, GA, 30322, USA \\
  \texttt{aawang6@emory.edu} \\\And 
  Susanna Spooner \\
  Department of Quantitative Science\\
  Emory University \\
  Atlanta, GA, 30322, USA \\
  \texttt{saspoon@emory.edu} \\}

\begin{document}
\pagenumbering{arabic}
\maketitle

\pagenumbering{arabic}
\begin{abstract}
\pagenumbering{arabic}
This study investigates game-based learning in the context of the educational game "Jo Wilder and the Capitol Case," focusing on predicting student performance using various machine learning models, including K-Nearest Neighbors (KNN), Multi-Layer Perceptron (MLP), and Random Forest. The research aims to identify the features most predictive of student performance and correct question answering. By leveraging gameplay data, we establish complete benchmarks for these models and explore the importance of applying proper data aggregation methods. By compressing all numeric data to min/max/mean/sum and categorical data to first, last, count, and nunique, we reduced the size of the original training data from 4.6 GB to 48 MB of preprocessed training data, maintaining high F1 scores and accuracy.

Our findings suggest that proper preprocessing techniques can be vital in enhancing the performance of non-deep-learning-based models. The MLP model outperformed the current state-of-the-art French Touch model, achieving an F-1 score of 0.83 and an accuracy of 0.74, suggesting its suitability for this dataset. Future research should explore using larger datasets, other preprocessing techniques, more advanced deep learning techniques, and real-world applications to provide personalized learning recommendations to students based on their predicted performance. This paper contributes to the understanding of game-based learning and provides insights into optimizing educational game experiences for improved student outcomes and skill development.

\end{abstract}
\section{Introduction}
\label{sec:introduction}

A radical shift that is currently transforming the field of education is the adoption of technology to facilitate student learning and growth in school; widespread usage of tablets, laptops, and mobile devices in primary and secondary education enables a universe of possibilities in terms of leveraging digital technology to enhance student learning \cite{schindler-2023-engagement}.  

In particular, game-based learning is an educational intervention that has received extensive interest in the field of pedagogy, for three key reasons. First, it has been proposed to increase student engagement, given that the elements of discovery, immersion, and feedback inherent to game-based environments can pique students' curiosity and motivate them to master the accompanying educational material. Existing evidence demonstrates that students in primary and secondary school exhibit lower learning anxiety when playing educational games instead of listening to traditional lectures, in addition to an increase in both motivation and academic performance within the context of several STEM disciplines \cite{hung-2014-digitallearning, ciloglu-2023-ARbiology}. Second, a key advantage of game-based learning and digital approaches in general is the increased accessibility and scale; akin to online platforms like Coursera and Udemy, these interventions could address educational inequalities and widen access to millions of students globally, particularly those in low-income homes or developing countries \cite{haleem-2022-digital}. Lastly, game-based learning offers the possibility of personalized approaches, such that the games are tailored and adapted to the skill level of each student. Thus, in spite of the fundamentally challenging and time-consuming task of developing educational games that are in alignment with current pedagogical guidelines and that cover different academic disciplines, work in the domain of game-based learning is crucial given its ability to promote student outcomes and skill development.

However, for game-based learning approaches to be effective, the underlying engine must accurately predict a student's performance throughout the game; otherwise,  feedback (e.g., hints, suggestions) and in-game checkpoints will not be implemented appropriately, such that students remain engaged with the material and are continually improving as they learn from the game. Despite significant advances in the development of several game-based frameworks \cite{min-2017-multimodal} and theoretical student learning models \cite{mayer-2019-games}, research on knowledge tracing has lagged behind, primarily due to the lack of large-scale, open-source datasets from game-based learning platforms that have amassed a significant population of student users. Furthermore, though knowledge tracing and characterization of a student's mastery level have been investigated in intelligent tutoring systems \cite{haridas-KT}, there has been little work focused on evaluating what specific features of a game-based learning environment support effective learning experiences for student; this is crucial given its implications in question design and pedagogical guidelines, in addition to the idea that we can only iterate and improve on these game-based approaches if we know what parts of the gameplay experience are driving the most and least learning.

In order to address the paucity of available evidence in these areas, we leverage student gameplay data from \textit{Jo Wilder and the Capitol Caese}, a popular educational game designed for students in elementary school and focused on the skills of historical analysis and reading comprehension; the dataset was released to encourage efforts to improve predictive student modeling and knowledge tracing (discussed above), which will ultimately help improve the gameplay design and enhance learning outcomes \cite{wardrip2022jo}. We hypothesize that by leveraging one of the only existing compiled databases of student gameplay interactions in a historical inquiry game, we can better determine the features that are most predictive of student performance and correct question answering. In addition, our empirical findings establish complete benchmarks in terms of several machine learning models, which will be discussed in Section \ref{sec:approach}. In summary, this paper makes three main contributions as follows:
\begin{enumerate}
    \item We leverage an open-access dataset from the \textit{Jo Wilder} online educational game, consisting of 23.6 million user interactions segmented across multiple game sessions, questions, and levels designed for 3rd to 5th grade history class students; we release the code and data for reproducibility at \url{https://github.com/HarryJeon24/Student_Performance}.
    \item We determine off-the-shelf, benchmark performance for Random Forest, K-Nearest Neighbors (KNN), and Multilayer Perceptron (MLP) models with standard preprocessing criteria to discern baseline scores in terms of accuracy and F1 scores; we achieve the state-of-the-art performance on the \textit{Jo Wilder} dataset with a F1 score of 0.83 using the MLP model, which is higher than the current best contender with a F1 score of 0.72.
    \item We establish benchmark performance for the multilayer perceptron, which is the first neural network model to be run on the dataset; details on parameters and performance will be discussed in Section \ref{sec:experiments}.
\end{enumerate}

This work will facilitate the further development of game-based learning approaches, especially in the context of question and gameplay design; to the best of our knowledge, this is the first published work on the \textit{Jo Wilder} educational game dataset using state-of-the-art machine learning models for student knowledge tracing.

\section{Background and Related Work}
\label{sec:related-work}

Recent work on evaluating game-based learning approaches has spanned a variety of domains in STEM and the humanities, including but not limited to biology, mathematics, and reading comprehension \cite{hung-2014-digitallearning, min-2017-multimodal, ciloglu-2023-ARbiology}. However, most of the existing research has taken the perspective of investigational studies, focused on interviewing students and understanding factors such as their motivation, engagement, and performance while playing different educational games; this is in contrast to directly modeling student knowledge (i.e., knowledge tracing) and leveraging computational approaches to determine how much a student has mastered the material.

\citet{chu-concept-mathematics} improved on these qualitative approaches by introducing quantitative metrics to represent a student's comprehension for different mathematical concepts in a primary school educational game, including competencies such as the multiplication of integers and fractions. By scoring each question in the game with a relevancy index (i.e., to explictly label what learning objective each question was designed to address), the in-game checkpoint assessment scores were used to decide which of three alternative learning paths was presented to the students; thus, this allowed the underlying game engine to adopt a diagnostic approach, such that students who have not yet attained the expected learning outcomes are instructed with further supplementary content. The post-gameplay scores demonstrated that students on average scored approximately 13\% higher than students who listened to traditional lecture \cite{chu-concept-mathematics}. Despite these advances, however, it is evident that multiple layers of information were left out from consideration, including but not limited to how much time was spent in each game session, what text the students saw, which questions they spent the most time on, where they placed their cursor, and what coordinates in the game they clicked on.

In particular, \citet{geden_emerson_carpenter_rowe_azevedo_lester_2020} utilized similar time-stamped gampelay features (in-game assessment scores, level of the game that a student is on, what the students are clicking and which gameplay actions they engage in) to predict student microbiology content knowledge, achieving the optimal F1-score of 0.65 with a linear support vector machine (SVM) and 0.63 with k-nearest neighbors (KNN). These features are similar to those that are available to us with the \textit{Jo Wilder} dataset, which will be further explored in Section \ref{sec:approach}; however, since this dataset is novel, we are unable to disclose any comparative metrics. To the best of our knowledge, machine learning approaches to model student mastery has not been performed on a historical inquiry game of this scale.

Furthermore, it is important to note that several pioneering studies have begun to incorporate multimodal data to augment model performance in terms of knowledge tracing; \citet{emerson-multimodal} took advantage of not only student gameplay interactions, but also data on facial expression and eye gaze captured by camera sensors on the computer workstations students were using. Applying a logistic regression model, a F1-score of 0.607 was achieved, demonstrating that complex multimodal, data inputs may not necessarily translate to better knowledge tracing as compared to previous work by \citet{geden_emerson_carpenter_rowe_azevedo_lester_2020}. As such, in the context of this study, we decide to make use of only gameplay interactions, and address gaps in the literature by examining whether certain features from student in-game interaction data from a historical inquiry game disproportionately account for accurate predictions on whether a student correctly answers an in-game assessment question or not (i.e., as a proxy for student mastery of the material). Our work is distinguished because we propose to evaluate the existing dataset with a neural network model (i.e.,  multilayer perceptron), which has not been experimented before to the best of our knowledge.

\section{Approach/Method}
\label{sec:approach}

\subsection{Machine Learning Techniques}

We trained K-Nearest Neighbors (KNN), Multilayered Perceptron, and Random Forest models on our data. 

KNN is a commonly used machine learning classifier that is useful because it does not make assumptions about the underlying distribution of the data. In KNN, the training data is saved, and, when presented with a new observation to classify, the points that are most similar to the new observation are used to classify it by selecting the most common label of the points.\cite{KNN-Pouriyeh-2023} The similarity of two points is calculated using a distance function such as Euclidean distance, Manhattan distance, or cosine similarity. This function is chosen at the time of training the KNN model. Euclidean Distance is defined as:
$$D(x, y) = \sqrt{\sum_{i = 1}^m (x_i - y_i)^2}$$
Manhattan Distance is defined as:
$$D(x, y) = \sum_{i = 1}^m |x_i - y_i|$$
Cosine similarity is defined as:
\begin{equation} 
    \begin{split}
        similarity & = cos(\theta)  \\
        & = \frac{A \cdot B}{\|A\|\|B\|} \\
        & = \frac{ \sum_{i=1}^{n} A_iB_i }{ \sqrt{ \sum_{i=1}^nA_i^2 } \sqrt{ \sum_{i=1}^n B_i^2 }}
    \end{split}
\end{equation}

The number of most similar points to use is represented by the hyperparameter, k. One of KNN's main drawbacks is that it suffers from the curse of dimensionality where as the number of features increases, the amount of observations needed to obtain a good accuracy exponentially increases.\cite{KNN-Pouriyeh-2023} In this case, this is not an issue as our dataset is large and we are not using that many features. Another drawback of KNN is that it is computationally expensive as the model needs to store all of the training data and calculate each distance. However the simplicity of this machine learning algorithm and the fact that it does not make assumptions about the data distributions is incredibly useful and make it a good fit for our data.

\begin{algorithm}[h]
\SetAlgoLined
\KwIn{Training data $(X, Y)$, test data $X_{test}$, number of neighbors $K$}
\KwOut{Predicted class labels for test data $Y_{test}$}
\For{$i = 1$ to $N_{test}$}{
  Calculate the distance $d(X_{test}^{(i)}, X^{(j)})$ between $X_{test}^{(i)}$ and all training data points $X^{(j)}$\;
  Sort the distances in ascending order and select the $K$ nearest neighbors\;
  Assign the majority class label of the $K$ neighbors as the predicted class label for $X_{test}^{(i)}$\;
}
\caption{K-Nearest Neighbors Algorithm}
\label{alg:knn}
\end{algorithm}

Multilayered Perceptron is a machine learning technique that uses multiple layers of perceptrons that are fully connected in an artificial neural network. In a single perceptron, inputs (features) are assigned weights. The weighted sum of the features will allow the classifier to predict if an observation is positive or negative based on its comparison to a threshold. The weights and biases are updated iteratively by the model's performance on each data point to minimize 0-1 loss. The equation is shown below:
$$ a = \phi (\sum_j w_j x_j + b)$$
where a is the perceptron's activation, $x_j$ are the perceptron's inputs or the features of an observation, $w_j$ are the weights, and b is the bias. We are summing over all of the inputs (features) so $w_j$ represents the weight for the jth feature.\cite{Grosse-2017} $\phi$ is the activation function which can be set in the training phase. It can take be set to a linear classifier or a logistic classifier for example.

With a single perceptron, the data must be linearly separable, but with multiple layers, the data does not need to be linearly separable. \cite{Grosse-2017} Multilayered Perceptron is a feed-forward neural network which means that the perceptron nodes are arranged in such a way that there are no cycles and a later layer will not feed into a previous layer. \cite{ScienceDirect-MultilayerPerceptron} The inputs of the next layer are the outputs of the previous layer. Thus if we have layer 1:
$$ a_1 = \phi (\sum_j w_j x_j + b) $$
(where $a_1$, $w_j$, and $b$ are vectors whose length is number of units, or width, of the layer). The output, $a_1$, feeds into layer 2:
$$ a_2 = \phi (\sum_j w_j a_2 + b) $$
and so on. 

One of the drawbacks of neural networks is that the results are not interpretable and the model does not produce information about the correlations of the features to the label or the feature importance. \cite{Grosse-2017} This algorithm's strengths lie in that fact that it is able to work with with complex, non-linear problems and it benefits from large inputs of data. As we have a lot of data and a rather complicated problem, this model is a suited for our data. \cite{ResearchGate-MLP-Advantages-Disadvantages}

\begin{algorithm}[h!]
\SetAlgoLined
\KwIn{Training data $(X, Y)$}
\KwOut{Trained model parameters $\theta$}
Initialize weights and biases randomly\;
\While{not converged}{
  Sample a minibatch $(X_{batch}, Y_{batch})$ from $(X, Y)$\;
  Compute the forward pass:
  \begin{align*}
    Z_1 &= X_{batch}W_1 + b_1 \\
    A_1 &= \sigma(Z_1) \\
    Z_2 &= A_1W_2 + b_2 \\
    A_2 &= \sigma(Z_2) \\
    \vdots \\
    Z_L &= A_{L-1}W_L + b_L \\
    A_L &= \sigma(Z_L) \\
    \hat{Y}_{batch} &= A_L
  \end{align*}
  Compute the loss $L(\hat{Y}_{batch}, Y_{batch})$\;
  Compute the gradients with respect to the loss:
  \begin{align*}
    \frac{\partial L}{\partial \hat{Y}_{batch}} &= \nabla_{\hat{Y}_{batch}} L \\
    \frac{\partial L}{\partial Z_L} &= \frac{\partial L}{\partial \hat{Y}_{batch}} \odot \sigma'(Z_L) \\
    \vdots \\
    \frac{\partial L}{\partial Z_1} &= \frac{\partial L}{\partial Z_2}W_2^\top \odot \sigma'(Z_1)
  \end{align*}
  Update the weights and biases using the gradients:
  \begin{align*}
    W_1 &\leftarrow W_1 - \alpha \frac{\partial L}{\partial W_1} \\
    b_1 &\leftarrow b_1 - \alpha \frac{\partial L}{\partial b_1} \\
    \vdots \\
    W_L &\leftarrow W_L - \alpha \frac{\partial L}{\partial W_L} \\
    b_L &\leftarrow b_L - \alpha \frac{\partial L}{\partial b_L}
  \end{align*}
}
\caption{Multilayer Perceptron Training Algorithm}
\end{algorithm}

The third machine learning technique that we used is Random Forest. In Random Forest, multiple decision trees are modeled on the data in order to produce multiple predictions which are combined together in a majority vote to produce one prediction. 

In a decision tree, the data is partitioned using hyperrectangles. Decision trees take the dataset, select the "best" feature, and split the data into subsets (based on a threshold value or categories). There are two commonly used ways of finding the "best" feature to split the data on: information gain using entropy and Gini impurity. \cite{DT-Pouriyeh-2023} Entropy is calculated as:

$$\text{Entropy} = -\sum_{i=1}^n p_ilog_2p_i$$
Where $p_i$ is the probability of randomly selecting an example in class $i$.
Gain is calculated as:
$$\text{Gain(S, A)} = \text{Entropy(S)}-\sum_{v \in Values(A)} \frac{|S_v|}{|S|} $$
Where, Values(A) is all the possible values for attribute A, and $S_v$ is the subset for S for which attribute A has value $v$. 
Gini Impurity is calculated as:
$$\text{Gini Impurity} = 1-\sum_{i = 1}^n p_i^2 $$ \cite{DT-Pouriyeh-2023}

The select and split process is repeated on each of the subsets until all the instances a subset has the same class, there are no more features left, the maximum depth has been reached, or the minimum number of observations in the subset has been reached. The nodes that are not partitioned further (leaf nodes) represent the hypercubes. They are given a label by taking the most frequent label among the observations in the node (majority vote). \cite{DT-Pouriyeh-2023}

Random Forest constructs many trees by using bootstrapping aggregation, or bagging. In other words, a random subset of the training data is selected with replacement to build each tree. \cite{RF-McAlister-2023} Thus, each tree in the forest will produce a prediction result, and the label that is predicted the most (majority vote) is selected. This method has the benefits of reducing overfitting, providing flexibility, and being able to evaluate feature importance. \cite{DT-Pouriyeh-2023} Being able to access feature importance could be quite useful for educational game developers seeking to improve how much their players learn and how well they can answer questions about the game's topic. Additionally, reducing overfitting is useful as our data contains much noise and it would lower the model's ability to classify new observations if the model fit too closely to the noise.

\begin{algorithm}[h]
\SetAlgoLined
\KwIn{Training data $(X, Y)$, number of trees $T$, maximum depth $D$}
\KwOut{Random Forest model $\Theta$}
\For{$t = 1$ to $T$}{
Sample a bootstrap dataset $(X_t, Y_t)$ from $(X, Y)$;
Train a decision tree $f_t$ on $(X_t, Y_t)$ with maximum depth $D$;
}
\KwRet{Ensemble model $\Theta = {f_1, f_2, ..., f_T}$}
\caption{Random Forest Algorithm}
\label{alg:rf}
\end{algorithm}

\section{Experiments}
\label{sec:experiments}
    
\subsection{Data Exploration}
As stated in the introduction, our data comes from students interacting with the \textit{Jo Wilder and the Capitol Caese} game. The data consists of 23.6 million user interactions which includes clicking on multiple types of objects, hovering over maps or objects, and reaching checkpoints. Each observation comes from one of 23.5 thousand user sessions. Each session is split into 23 levels (0 to 22). At the end of each level, there is a question. Data from the first 18 questions were provided in a separate data set. We have 424 thousand observations. Almost 300 thousand of the questions were answered correctly, and the other 124 thousand were answered incorrectly. 

Features:
\begin{itemize}
    \item session\_id - the ID of the session in which the event takes place
    \item index - the index of the event
    \item elapsed\_time - the amount of time (in milliseconds) between the start of the session and the event
    \item event\_name - the name of the event type
    \item name - the name of the event
    \item level - the level that the event took place in (0 to 22)
    \item page - the page number of the event in notebook events
    \item room\_coor\_x - the x-coordinate of the click in the in-game room
    \item room\_coor\_y - the y-coordinate of the click in the in-game room
    \item screen\_coor\_x - the x-coordinate of the click on the player’s screen
    \item screen\_coor\_y - the y-coordinate of the click on the player’s screen
    \item hover\_duration - duration of hover events in (milliseconds)
    \item text - the text the player sees during this event
    \item fqid - the fully qualified ID of the event
    \item room\_fqid - the fully qualified ID of the room the event took place in
    \item text\_fqid - the fully qualified ID of the text that the player sees
    \item fullscreen - 1 if the user played in fullscreen mode, 0 otherwise
    \item hq - 1 if the user played the game in high quality, 0 otherwise
    \item music - 1 if the user played the game with music, 0 otherwise
    \item level\_group - combines levels and questions into groups via bins (0-4, 5-12, 13-22)
\end{itemize}

\subsection{Pre-Processing}

\subsubsection{Feature Extraction}
As we had many more observations than we did labels, we needed to aggregate the data for each level. We tried many methods in order to find the best method for each feature. For numerical features, we tried mean, sum, min, and max for aggregation. For categorical features, we tried first, last, count and nunique.

\subsubsection{Feature Selection}

We decided to drop columns that were missing large amounts of data. Thus, we dropped 
\textit{page}, \textit{hover\_duration} and \textit{text\_fqid}. 

Additionally, we decided to drop the \textit{text} column that contains natural language as its values are unstructured data and not particularly useful in this case since they consist of character dialogue that the game generates, and do not add additional information about the player's interactions with the game.

We use Pearson correlation coefficients in order to choose which features to include in our data (see figure 1). 

Pearson correlation for the population is calculated as:
$$ \rho_{X,Y} = \frac{cov(X, Y)}{\sigma_X\sigma_Y}$$
and Pearson correlation for samples is calculated as:

$$ r_{xy} = \frac{\sum_{i=1}^n(x_i - \Bar{x}(y_i - \Bar{y}}{\sqrt{\sum_{i=1}^n(x_i - \Bar{x})^2} \sqrt{\sum_{i=1}^n(y_i - \Bar{y})^2}}$$ \cite{Feat-Pouriyeh-2023}

The result is a number from [-1, 1] where 1 represents a perfect positive correlation, -1 represents a perfect negative correlation, and 0 represents completely uncorrelated variables. Thus, larger absolute correlation values between the label feature and another feature indicates that the feature is more important.  \cite{Feat-Pouriyeh-2023} In figure 1, the rightmost column and bottom-most row represent the label and its correlation to the other features. None of the correlations are particularly large aside from the correct label being correlated to itself (which makes sense). However, there are a few features such as fqid\_count that have slightly larger (absolute) values compared to the rest.

We selected the aggregation method that correlated the most with our label (whether the student answered the question correctly or not) and made sure that none of the features that we kept were correlated with each other. 

\begin{figure*}
  \includegraphics[scale = 0.4]{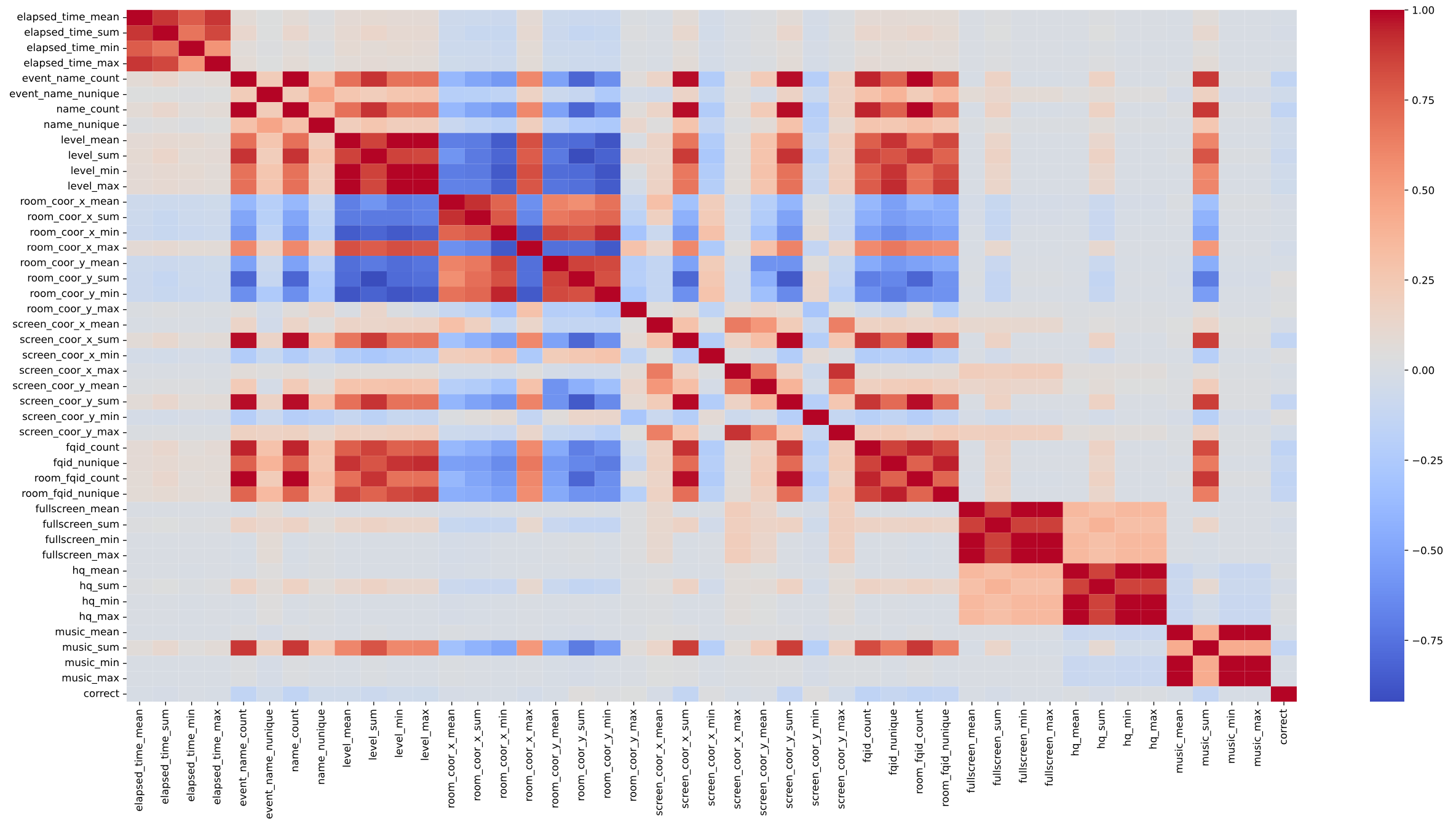}
  \caption{Pearson Correlation Matrix Heatmap.}
\end{figure*}

Additionally, we used information gain as another method of checking which of our variables are most important. We calculated the mutual information criterion (information gain) between all of the features and the label. Mutual information measures the dependency between two variables, or how closely related two variables are, so the features that have higher mutual information scores are more important to our model compared to features that have lower scores that are close to 0. A feature score of 0 indicates that the features are independent of each other. \cite{scikit-learn-mutual-info-classif}

To calculate information gain, the following equation is used:
$$I(X; Y) = \sum_{x,y}p(x,y)log\frac{p(x,y)}{p(x)p(y)}$$  \cite{Feat-Pouriyeh-2023}

The results of the information gain are shown below:
\begin{figure}
  \includegraphics[scale = 0.4]{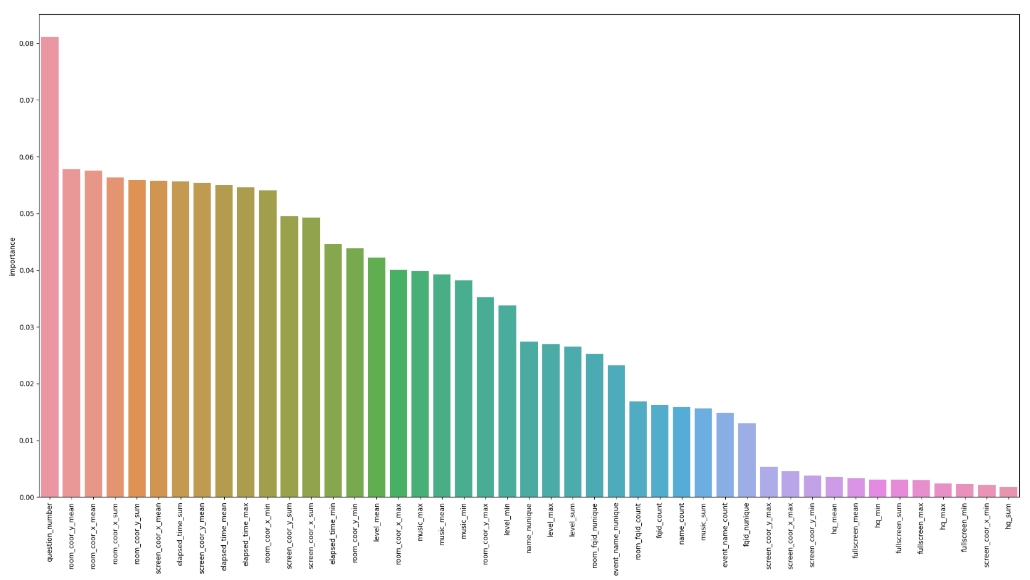}
  \caption{Bar plot of importance.}
\end{figure}

After examining the Pearson correlation coefficients and information gain (i.e., importance) values, we decided to use the following features based on the highest possible reductions in entropy; the corresponding information gain values are included for reference:

\begin{itemize}
    \item room\_coor\_x\_mean: the mean coordinate value of the x clicks in the in-game room; information gain value $\approx 0.0567$
    \item room\_coor\_y\_mean: the mean coordinate value of the y clicks in the in-game room; information gain value $\approx 0.0573$
    \item screen\_coor\_x\_mean: the mean coordinate value of the x clicks on the computer screen; information gain value $\approx 0.0559$
    \item screen\_coor\_y\_mean: the mean coordinate value of the y clicks on the computer screen; information gain value $\approx 0.0553$
    \item elapsed\_time\_sum: the sum of the amount of time between the session start and the events; information gain value $\approx 0.0558$
    \item level\_mean: the average of the levels; information gain value $\approx 0.0431$
    \item music\_max: 1 if the music was on at all in the level, 0 otherwise; information gain value $\approx 0.0394$
    \item name\_nunique: the number of unique event names; information gain value $\approx 0.0269$
    \item room\_fqid\_nunique: the number of unique room IDs; information gain value $\approx 0.0268$
    \item event\_name\_nunique: the number of unique event types; information gain value $\approx 0.0249$
    \item fqid\_count: the number of IDs; information gain value $\approx 0.0175$
\end{itemize}

\subsubsection{Modeling Choices}
In this study, we selected three distinct machine learning models to predict whether a student correctly solved a problem based on their performance in gameplay. These models were chosen due to their ability to capture non-linear patterns in the dataset, as identified by the correlation matrix and information gain graph analyses. The selected models are:

\begin{enumerate}
\item \textbf{Random Forest}: An ensemble learning technique that constructs multiple decision trees and aggregates their predictions, offering improved accuracy and reduced overfitting. This method is well-suited for handling linear and non-linear relationships between features and the target variable.
\item \textbf{k-Nearest Neighbors (KNN)}: A straightforward, non-parametric algorithm applicable for both classification and regression tasks. KNN computes the distance between a test point and its k-nearest neighbors in the training set, assigning the most frequent class among the neighbors.
\item \textbf{Multi-Layer Perceptron (MLP)}: A type of artificial neural network comprising one or more hidden layers, which can effectively capture complex, non-linear relationships between input features and the target variable.
\end{enumerate}

To ensure a reliable evaluation of these models, we employed k-fold cross-validation with 5 folds for Random Forest and MLP, and 10 folds for KNN. This process involved dividing the dataset into k equal parts, training the model on k-1 parts, and testing it on the remaining part, iterating k times. We calculated the accuracy and F1-score for each fold, averaging these scores to obtain a single performance metric for each model.

Parameters for each model were determined as follows:

\begin{itemize}
\item \textbf{Random Forest}: We employed 100 estimators (decision trees) and set the random state to 42.
\item \textbf{KNN}: We opted for 5 neighbors (k) and preprocessed the data using StandardScaler for numerical features and OneHotEncoder for categorical features before fitting the KNN model.
\item \textbf{MLP}: The input dimension was set equal to the number of features in the dataset, the hidden layer size to 128, and the output dimension to 2 (assuming binary classification). The neural network was trained for 100 epochs using the Adam optimizer with a learning rate of 0.001 and the CrossEntropyLoss as the loss function.
\end{itemize}

In our analysis, we addressed the potential class imbalance issue by aggregating the data, which resulted in a more balanced dataset. Furthermore, we handled missing values using the mean imputation strategy and normalized the data to improve its quality. These preprocessing steps contributed to a more balanced class distribution, eliminating the need for specific class balancing techniques such as subsampling, oversampling, or alternative methods.

Although we did not incorporate subsampling, oversampling, or other class-balancing techniques in our analysis, it is important to note that these methods can be considered in cases where class imbalance persists despite aggregation and preprocessing. In such scenarios, implementing strategies like random undersampling, random oversampling, or the Synthetic Minority Over-sampling Technique (SMOTE) can help to balance the dataset and potentially enhance the model's performance. However, for our specific dataset and problem, the aggregation and preprocessing steps effectively addressed the class imbalance issue, making the use of additional balancing techniques unnecessary.

\subsubsection{Empirical Results and Comparisons}

Table \ref{tab:results} shows the F-1 score and accuracy for four different models.

\begin{table}[htbp]
  \centering
  \caption{F-1 score and accuracy for 4 models}
    \begin{tabular}{cccc}
    \toprule
    \textbf{Model} & \textbf{F-1 Score} & \textbf{Accuracy} \\
    \midrule
    French Touch & 0.72 & N/A \\
    Random Forest & 0.65 & 0.73 \\
    KNN & 0.56 & 0.70 \\
    \textbf{MLP} & \textbf{0.83} & \textbf{0.74} \\
    \bottomrule
    \end{tabular}%
  \label{tab:results}%
\end{table}%

As presented in Table \ref{tab:results}, the French Touch model, which is considered the current state-of-the-art in predicting student problem-solving performance in gameplay, achieved an F-1 score of 0.72. While the accuracy metric was not available for the French Touch model, its F-1 score outperformed both the Random Forest and KNN models. This demonstrates the efficacy of the French Touch model in handling complex relationships and capturing nuances in the data.

However, the MLP model surpassed the French Touch model in terms of both F-1 score (0.83) and accuracy (0.74), suggesting that the MLP model may be more suitable for this particular dataset. The impressive performance of the MLP model can be attributed to its ability to capture complex, non-linear relationships between input features and the target variable.
\section{Discussion}
\label{sec:analysis}


\subsection{Pre-Processing}

Our experimental methods and results have highlighted the importance of applying proper data aggregation methods for training on a large dataset. Specifically, by compressing all numeric data to min/max/mean/sum and all categorical data to first/last/count/nunique, we were able to reduce the size of the original training data from 4.6 GB to 48 MB of preprocessed training data. While this process resulted in a loss of information, we obtained a 1D feature set to 1 label trainable data, which our models could read in while maintaining a very high F1 score and accuracy. Additionally, we dropped feature columns that had little relation to other features or provided little information gain to resolve some of the overfitting issues and further improve our model's performance.

Our findings suggest that proper preprocessing techniques can be vital in enhancing the performance of non-deep-learning-based models. In particular, our Random Forest model achieved an accuracy of 0.73 after preprocessing, which was only 0.01 below the 0.74 accuracy achieved by our MLP model. Given the complexity of the dataset, this performance gain is remarkable.

\subsection{Model Performances}

Our primary finding is the effectiveness of using deep learning models in predicting student game performance. Although our MLP model had similar accuracy to the Random Forest model and KNN model, the F1 score for MLP was significantly higher. This may be due to the non-regressional nature of the features in the dataset. For example, the preprocessed feature column count for event name has no proper threshold for the decision tree in the random forest model to determine. Similarly, the KNN model is unable to grasp the nuances between the features, such as differences in the type and quality of clicks. Our MLP model addressed these issues by fitting a multi-layer model that captured the complexity of the relationships between features and student performance.

\subsection{Future Directions}

Our study has identified several promising avenues for future research. First, in order to incorporate our research models into real-world applications, we suggest using a larger dataset to provide more generalizable results and a better understanding of the relationships between features and student performance. Additionally, we recommend exploring other preprocessing techniques, such as dimensionality reduction or feature engineering, to further improve model performance.

Another area for future research is to investigate the use of more advanced deep learning techniques, such as Convolutional Neural Networks (CNNs) or Recurrent Neural Networks (RNNs), which have been successful in other machine learning domains. These techniques may be particularly useful for capturing temporal relationships between student actions and performance in the game.

Finally, we propose applying our model to real-world scenarios and evaluating its feasibility in providing personalized learning recommendations to students based on their predicted performance. This could involve collaborating with educators to implement our model in an educational setting and evaluating its effectiveness in improving student outcomes. Overall, our study provides a promising starting point for further investigation into using machine learning to improve student performance in educational games.

\pagenumbering{arabic}
\section*{Contributions}

All members contributed to the pre-processing, feature extraction, feature selection, and model selection.
Harry Jeon ran the KNN and multi-layer perceptron models and helped to write experiments/results sections. Harry He trained the Random Forest models and wrote the discussion section. Anthony Wang wrote the background and introduction sections. Susanna Spooner wrote the experiments/results and methods sections.

\pagenumbering{arabic}

\bibliography{custom}


\end{document}